\documentclass[11pt,letterpaper]{article}
\usepackage{emnlp2017}
\usepackage{CJKutf8}
\emnlpfinalcopy

\def\emnlppaperid{***}

\usepackage{times}
\usepackage{latexsym}
\usepackage{amsmath}
\usepackage{amssymb}
\usepackage{multirow}
\usepackage{url}
\makeatletter

\usepackage{booktabs}
\usepackage{graphicx}

\title{Encoding Sentences with Graph Convolutional Networks \\ for Semantic Role Labeling}

 \author{
 Diego Marcheggiani$^1$ \hspace{1cm}  Ivan Titov$^{1,2}$    \\
 $^1$ILLC,  University of Amsterdam  \\
   $^2$ILCC, School of Informatics, University of Edinburgh \\
    {\tt marcheggiani@uva.nl} \hspace{1cm} {\tt ititov@inf.ed.ac.uk}
 }

\date{}

\begin{document}
\maketitle
\begin{abstract}
Semantic role labeling (SRL) is the task of identifying the predicate-argument structure of a sentence. 
It is typically regarded as an important step in the standard NLP pipeline.
As the semantic representations are closely related to syntactic ones, we exploit syntactic information in our model.  
We propose a version of graph convolutional networks (GCNs), a recent class of  
neural networks operating on graphs, suited to model syntactic dependency graphs. 
GCNs over syntactic dependency trees are used as sentence encoders, producing latent feature representations of words in a sentence. 
We observe that GCN layers are complementary to LSTM ones: when we stack both GCN and LSTM layers, we obtain a substantial improvement over an already state-of-the-art LSTM SRL model, resulting in the best reported scores on the standard benchmark (CoNLL-2009) both for Chinese and English.
\end{abstract}

\section{Introduction}

\begin{figure}
\begin{center}
\includegraphics[width=1 \columnwidth, height=3.65cm, keepaspectratio]{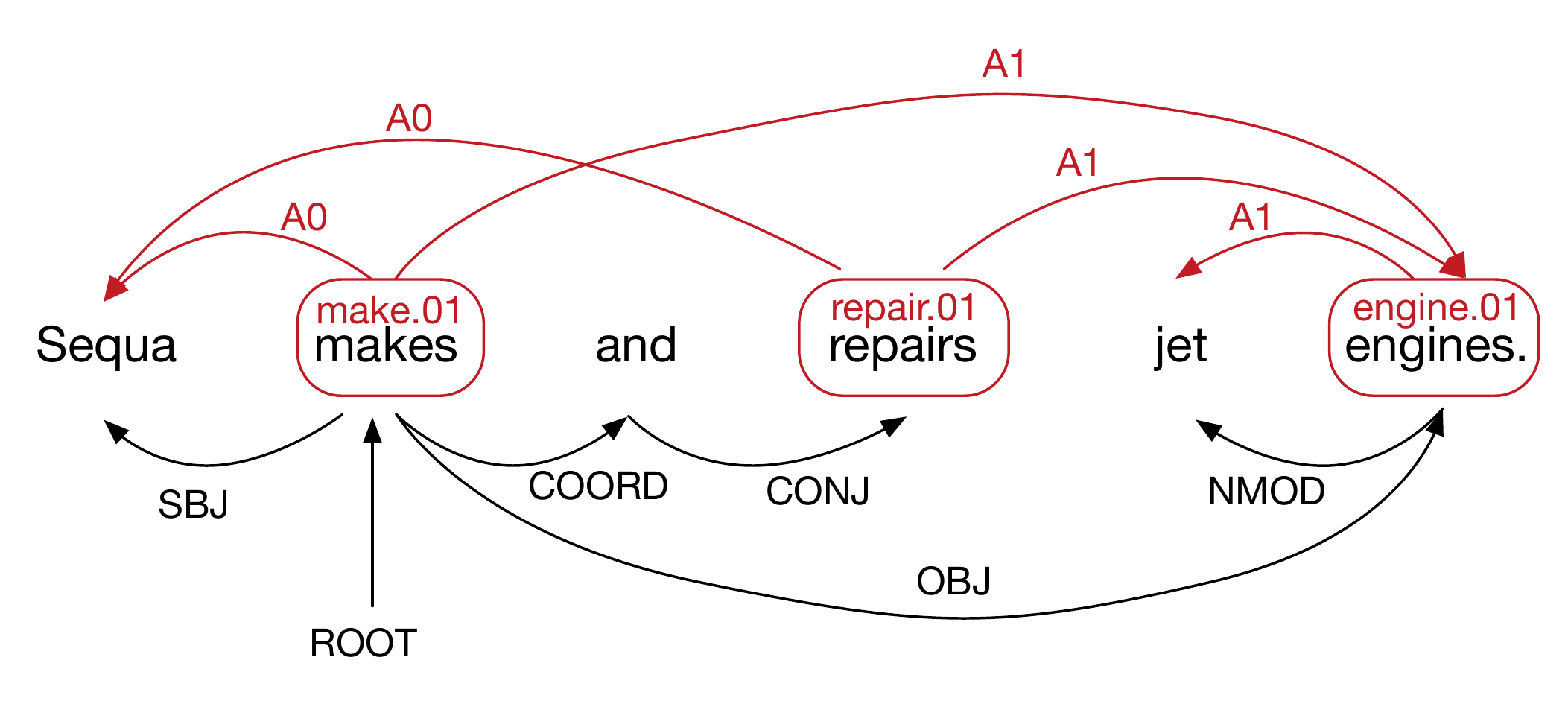}
\caption{ 
\label{fig:syntax-semantics} An example sentence annotated with semantic (top) and syntactic dependencies (bottom).
} 
\end{center}
\end{figure}

Semantic role labeling (SRL) \cite{gildea2002automatic} can be informally described as the task of discovering \textit{who} did \textit{what} to \textit{whom}. For example, consider an SRL dependency graph shown above the sentence  in Figure~\ref{fig:syntax-semantics}.  Formally, the task includes
(1) detection of predicates (e.g., {\it makes});  (2) labeling the predicates with a sense from a sense inventory (e.g., {\it make}.01);  (3) identifying and assigning arguments to {\it semantic roles} (e.g., {\it Sequa} is A0, i.e., an agent / `doer' for the corresponding predicate, and  {\it engines} is A1, i.e., a patient / `an affected entity').  SRL is often regarded as an important step in the standard NLP pipeline, providing information to downstream tasks such as information extraction and question answering. 

The semantic representations are closely related to syntactic ones, even though the syntax-semantics interface is far from trivial~\cite{levin1993english}. 
For example, one can observe that many  arcs in the syntactic dependency graph (shown in black below the sentence in Figure~\ref{fig:syntax-semantics}) are mirrored in the semantic dependency graph. Given these similarities and also because of availability of accurate syntactic parsers for many languages, it seems natural to exploit syntactic information when predicting semantics.   Though historically most SRL approaches did rely on syntax ~\cite{DBLP:conf/ecml/ThompsonLM03,DBLP:conf/conll/PradhanHWMJ05,DBLP:journals/coling/PunyakanokRY08,DBLP:conf/coling/JohanssonN08},
 the last generation of SRL models
put syntax aside in favor of neural sequence models, namely LSTMs \cite{zhou-xu:2015:ACL-IJCNLP,DBLP:journals/corr/MarcheggianiFT17}, and outperformed syntactically-driven methods on  standard benchmarks. 
We believe that one of the reasons for this radical choice
is the lack of simple and effective methods
for incorporating syntactic information into sequential neural networks (namely, at the level of words). 
In this paper we propose one way how to address this limitation.

Specifically, we rely on graph convolutional networks (GCNs)~\cite{NIPS2015_5954,DBLP:journals/corr/KipfW16,kearnes2016molecular}, a recent class of multilayer neural networks operating on graphs.   For every node in the graph (in our case a word in a sentence),  GCN encodes relevant information about its neighborhood as a real-valued feature vector. 
GCNs have been studied largely in the context of undirected unlabeled graphs.
We introduce a version of GCNs for modeling syntactic dependency structures and generally applicable to labeled directed graphs.

One layer GCN encodes only information about immediate neighbors and $K$ layers are needed to encode $K$-order neighborhoods (i.e., information about nodes at most $K$ hops aways). 
This contrasts with recurrent and recursive neural networks~\cite{DBLP:journals/cogsci/Elman90,socher-EtAl:2013:EMNLP} which, at least in theory, can capture statistical dependencies across unbounded paths in a trees or in a sequence. 
However, as we will further discuss in Section~\ref{sec:complement}, this is not a serious limitation when GCNs are used in combination with  encoders based on recurrent networks (LSTMs). 
 When we stack GCNs on top of LSTM layers, we obtain a substantial improvement over an already state-of-the-art LSTM SRL model, resulting in the best reported scores on the standard benchmark (CoNLL-2009), both for English and Chinese.\footnote{The code is available at \url{https://github.com/diegma/neural-dep-srl}.}

Interestingly, again unlike recursive neural networks, GCNs do not constrain the graph to be a tree. 
We believe that there are many applications in NLP, where GCN-based encoders of sentences or even documents can be used to incorporate knowledge about linguistic structures (e.g., representations of syntax, semantics or discourse). For example, GCNs can take as input combined syntactic-semantic graphs (e.g., the entire graph from Figure~\ref{fig:syntax-semantics}) and be used within downstream tasks such as machine translation or question answering. However, we leave this for future work and here solely focus on SRL.

The contributions of this paper can be summarized as follows:
\begin{itemize}
\item we are the first to show that GCNs are effective for NLP;
\item we propose a generalization of GCNs suited to encoding syntactic information at  word level;
\item we propose a GCN-based SRL model and obtain state-of-the-art results on English and Chinese portions of the CoNLL-2009 dataset;
\item we show that bidirectional LSTMs and syntax-based GCNs have complementary modeling power.

\end{itemize}

\section{Graph Convolutional Networks}\label{sec:gcns}

In this  section we describe GCNs of \newcite{DBLP:journals/corr/KipfW16}. Please refer to \newcite{gilmer2017neural} for a comprehensive overview of GCN versions.

GCNs are neural networks operating on graphs and 
inducing features of nodes (i.e., real-valued vectors /  embeddings)  based on properties of their  neighborhoods. In \newcite{DBLP:journals/corr/KipfW16}, they were shown to be very effective for the node classification task: the classifier was estimated jointly with a GCN, so that the induced node features were informative for the node classification problem. 
Depending on how many layers of convolution are used, GCNs can capture information only about immediate neighbors (with one layer of convolution) or any nodes at most $K$ hops aways (if $K$ layers are stacked on top of each other).

More formally, consider an undirected graph $\mathcal{G} =(\mathcal{V}, \mathcal{E})$, where  $\mathcal{V}$ ($|V|=n$) and $\mathcal{E}$ are sets of nodes and edges, respectively. \newcite{DBLP:journals/corr/KipfW16} assume that edges contain all the self-loops,
i.e., $(v, v) \in \mathcal{E}$  for any $v$.
We can define a matrix
$X \in \mathbb{R}^{m \times n}$ with each its column $x_v \in \mathbb{R}^m$ ($v \in \mathcal{V}$) encoding node features. The vectors can  either encode genuine features (e.g., this vector can encode the title of a paper if citation graphs are considered) or be a one-hot vector.
The node representation, encoding information about its immediate neighbors, is computed as
\begin{equation}
\label{eq:gcns}
h^{}_v =  ReLU\left(\sum_{u \in \mathcal{N}(v)}  ( W^{} x_{u}  + b) \right),
\end{equation}
where $W \in \mathbb{R}^{m \times m}$ and $b\in \mathbb{R}^{m}$ are  a weight matrix and a bias, respectively; $\mathcal{N}(v)$ are neighbors of $v$; 
$ReLU$ is the rectifier linear unit activation function.\footnote{We dropped normalization factors used in \newcite{DBLP:journals/corr/KipfW16}, as they are not used in our syntactic GCNs.}
Note that $v \in \mathcal{N}(v)$  (because of self-loops), so  the input feature representation of $v$  (i.e. $x_v$) affects its induced representation $h^{}_v$.

As in standard convolutional networks \cite{lecun-01a}, by stacking GCN layers one can incorporate higher degree neighborhoods:
\begin{equation}
\nonumber
h_v^{(k+1)} =  ReLU\left(\sum_{u \in \mathcal{N}(v)} W^{(k)} h^{(k)}_{u}  + b^{(k)} \right)
\end{equation}
where $k$ denotes the layer number and $h_{v}^{(1)} = x_v$.

\section{Syntactic GCNs}\label{sec:gcns-dep}

\begin{figure}[t!]
\begin{center}
\includegraphics[width= 0.90 \columnwidth]{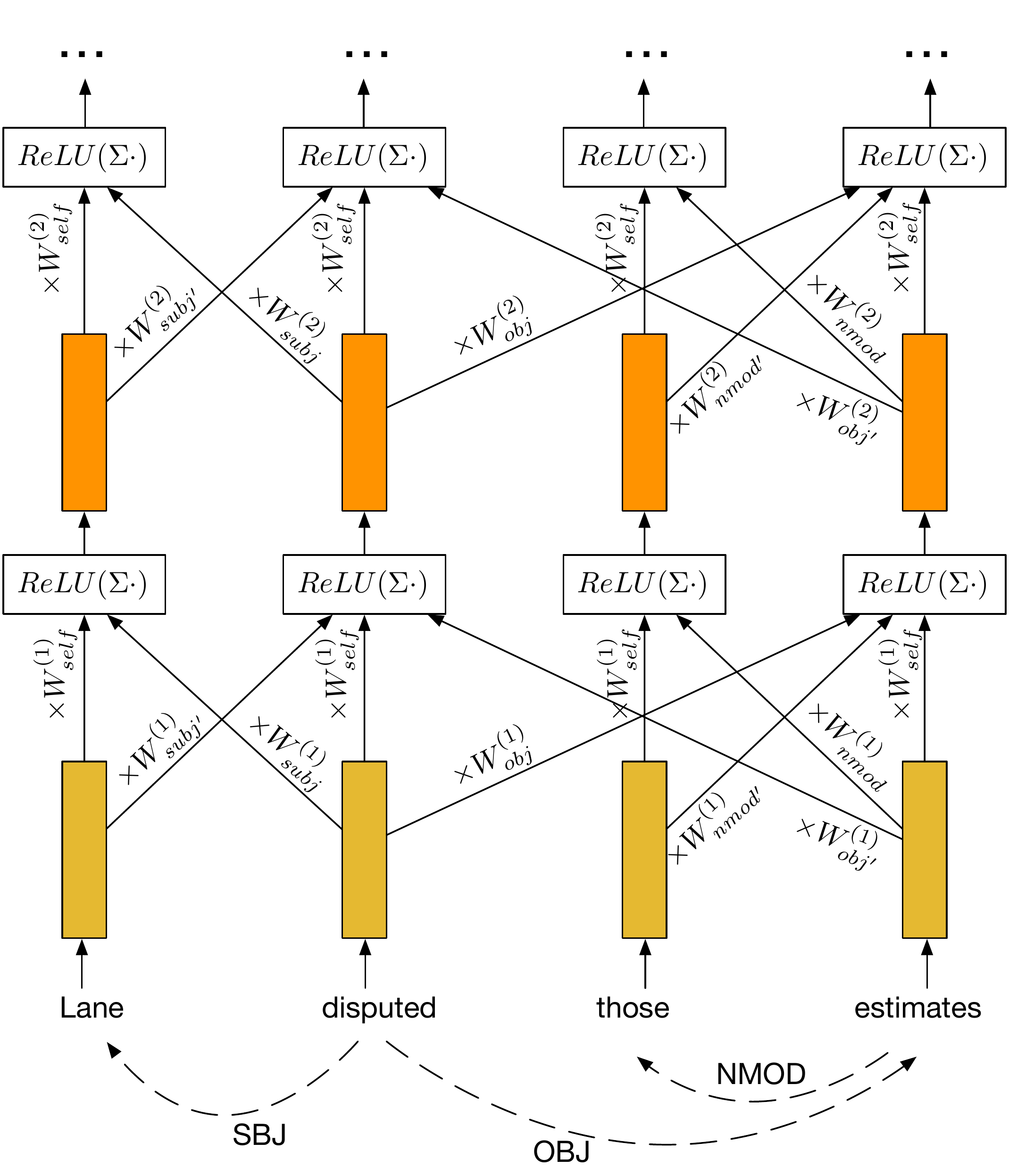}
\caption{\label{fig:syntax_gcn} A simplified syntactic GCN (bias terms  and gates are omitted); the syntactic graph of the sentence is shown with dashed lines at the bottom.
Parameter matrices are sub-indexed with syntactic functions, and apostrophes (e.g., {\it subj'}) signify  that  information flows in the direction opposite of the dependency arcs (i.e., from dependents to heads).}
\end{center}
\end{figure}

As syntactic dependency trees are directed and labeled (we refer to the dependency labels as {\it syntactic functions}), we first need to modify the computation in order to incorporate label information (Section~\ref{sec:incorporate}). In the subsequent section, we incorporate gates in GCNs, so that the model can decide which edges are more relevant to the task in question. Having gates is also important as we rely on automatically predicted syntactic representations, and the gates can  detect and downweight potentially erroneous edges. 

\subsection{Incorporating directions and labels}
\label{sec:incorporate}

Now, we introduce a generalization of GCNs appropriate for syntactic dependency trees, and in general, for directed labeled graphs.
First note that there is no reason to assume that information flows only along the syntactic dependency arcs (e.g., from {\it makes} to {\it Sequa}), so we allow it to flow in the opposite direction as well (i.e., from dependents to heads).
We use a graph $\mathcal{G} =(\mathcal{V}, \mathcal{E})$, where the edge set contains all pairs of nodes (i.e., words) adjacent in the dependency tree. 
In our example, both $(${\it Sequa}, {\it makes}$)$ and $(${\it makes}, {\it Sequa}$)$ belong to the edge set.
The graph is labeled, and the label $L(u,v)$ for $(u,v) \in \mathcal{E}$ contains both information about the syntactic function and indicates whether the edge is in the same or opposite direction as the syntactic dependency arc. 
For example, the label for $(${\it makes}, {\it Sequa}$)$ is $subj$, whereas the label for $(${\it Sequa}, {\it makes}$)$ is $subj'$, with the apostrophe indicating that the edge is in the direction opposite to the corresponding syntactic arc. 
Similarly, self-loops will have label $self$.
Consequently, we can simply assume that the GCN parameters are label-specific, resulting in the following computation, also illustrated in Figure~\ref{fig:syntax_gcn}:
\begin{equation}
\nonumber
h_v^{(k+1)} =  ReLU\left( \sum_{u \in \mathcal{N}(v)} W^{(k)}_{L(u,v)} h^{(k)}_{u}  + b^{(k)}_{L(u,v)} \right).
\end{equation}
This model is over-parameterized,\footnote{Chinese and English CoNLL-2009 datasets used 41 and 48 different syntactic functions, which would result in having 83 and 97 different matrices in every layer, respectively.}
especially given that  SRL datasets are moderately sized, by deep learning standards. So instead of learning the GCN parameters directly, we define them as
\begin{align}\label{eq:gcn-synt-compact} 
W^{(k)}_{L(u,v)} = V^{(k)}_{dir(u,v)}, 
\end{align}
where $dir(u,v)$ indicates whether the edge $(u,v)$ is directed (1) along, (2) in the opposite direction to the syntactic dependency arc, or (3) is a self-loop; $V^{(k)}_{dir(u,v)} \in \mathbb{R}^{m \times m}$. Our simplification captures the intuition that information should be propagated differently along edges depending whether this is a head-to-dependent or dependent-to-head edge (i.e., along or opposite the corresponding syntactic arc) and whether it is  a self-loop. So we do not share any parameters between these three very different edge types.  Syntactic functions are important, but perhaps less crucial, so they are encoded only in the feature vectors $b_{L(u,v)}$. 

\subsection{Edge-wise gating}
\label{sec:gating}

Uniformly accepting information from all neighboring nodes may not be appropriate for the SRL setting. For example, we see in Figure~\ref{fig:syntax-semantics} that many semantic arcs just mirror their syntactic counter-parts, so they may need to be up-weighted.
Moreover, we rely on automatically predicted syntactic structures, and, even for English, syntactic parsers are far from being perfect, especially when used out-of-domain. It is risky for a downstream application to rely on a potentially wrong syntactic edge, so the corresponding message in the neural network may need to be down-weighted. 

In order to address the above issues,
 inspired by recent literature \cite{DBLP:conf/nips/OordKEKVG16,DBLP:journals/corr/DauphinFAG16}, we calculate for each edge node pair a scalar gate of the form
\begin{equation}
g_{u, v}^{(k)} =  \sigma\left( h_u^{(k)} \cdot \hat{v}^{(k)}_{dir(u,v)} + \hat{b}^{(k)}_{L(u,v)}\right),
\end{equation}
where $\sigma$ is the logistic sigmoid function, $\hat{v}^{(k)}_{dir(u,v)} \in \mathbb{R}^{m}$ and $\hat{b}^{(k)}_{L(u,v)} \in \mathbb{R}$ are weights and a bias for the gate.
With this additional gating mechanism, the final syntactic GCN computation is formulated as
\begin{align}
\nonumber
h_v^{(k+1)} & \!\! =  \! ReLU\Large( \\ 
\label{eq:final-gcn}
& \!\!\sum_{u \in \mathcal{N}(v)}  \!\! g_{v, u}^{(k)} (V_{dir(u,v)}^{(k)} h^{(k)}_{u} \!  +  
b_{L(u,v)}^{(k)})\Large).
\end{align}

\subsection{Complementarity of GCNs and LSTMs}
\label{sec:complement}
The inability of GCNs to capture dependencies between nodes far away from each other in the graph
may seem like a serious problem, especially in the context of SRL: paths between predicates and arguments often include many dependency arcs~\cite{roth-lapata:2016:P16-1}. 
However, when graph convolution is performed on top of LSTM states (i.e., LSTM states serve as input ${x}_{v} = {h}_{v}^{(1)}$ to GCN) rather than static word embeddings, GCN may not need to capture more than a couple of hops.

To elaborate on this, let us speculate what role GCNs would play when used in combinations with LSTMs, given that LSTMs have already been shown very effective for SRL~\cite{zhou-xu:2015:ACL-IJCNLP,DBLP:journals/corr/MarcheggianiFT17}.
Though LSTMs are capable of capturing at least some degree of syntax~\cite{DBLP:journals/tacl/LinzenDG16} without explicit syntactic supervision, SRL datasets are moderately sized, so LSTM models may still struggle with harder cases. 
Typically, harder cases for SRL involve arguments far away from their predicates. In fact, 20\% and 30\% of arguments are more than 5 tokens away from their predicate, in our English and Chinese collections, respectively. 
However, if we imagine that we can `teleport' even over a single (longest) syntactic dependency edge, the 'distance' would shrink:
only  9\% and 13\%  arguments will now be more than 5 LSTM steps away (again for English and Chinese, respectively). 
GCNs provide this `teleportation' capability. These observations suggest that LSTMs and GCNs may be complementary, and we will see that empirical results support this intuition.

\section{Syntax-Aware Neural SRL Encoder}\label{sec:encoder}

\begin{figure}[t]
\begin{center}
\includegraphics[width=0.75 \columnwidth]{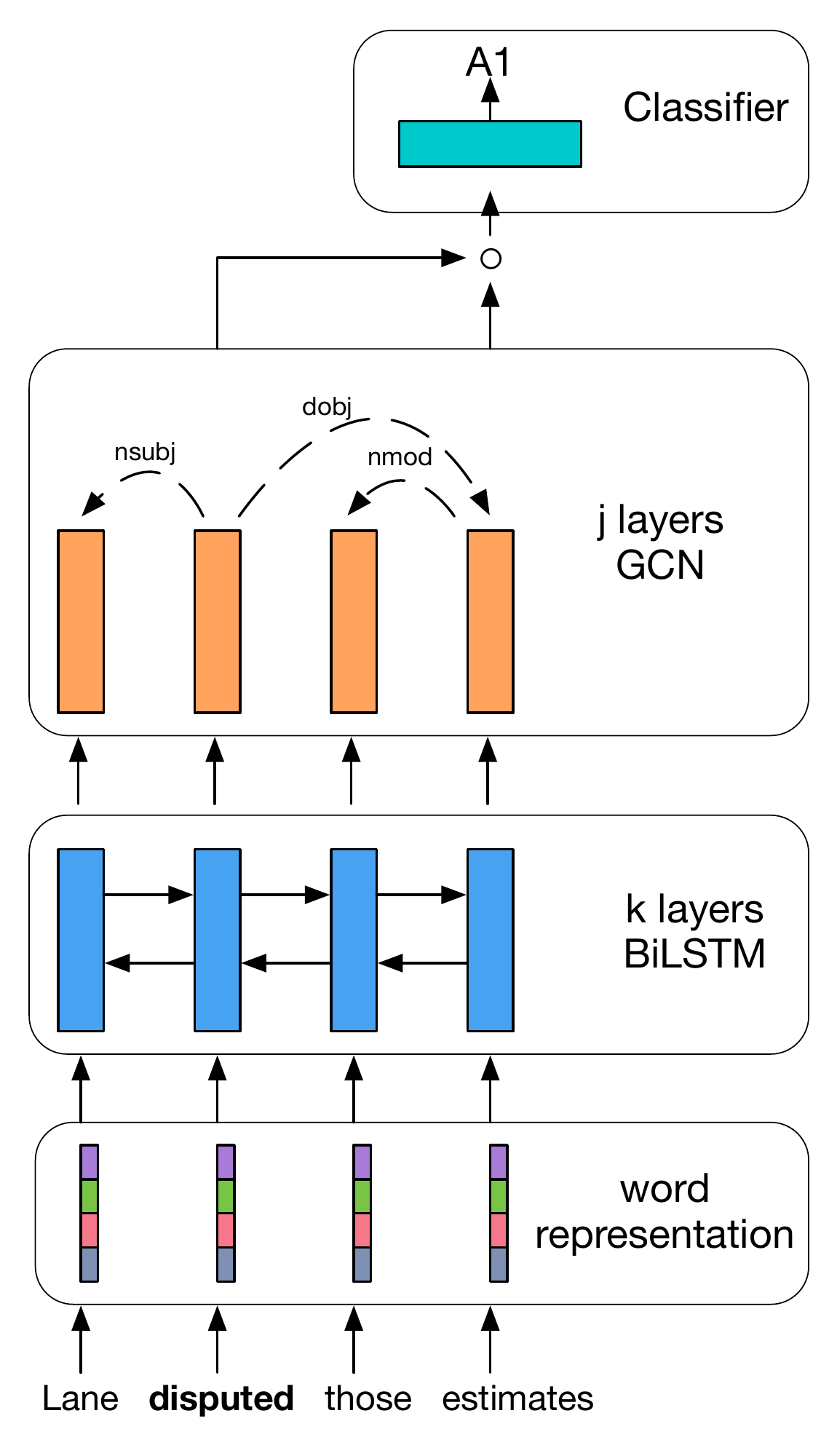}
\caption{\label{fig:model} Predicting an argument and its label with an LSTM + GCN encoder. 
} 
\end{center}
\end{figure}

In this work, we build our semantic role labeler on top of the syntax-agnostic LSTM-based SRL model of \newcite{DBLP:journals/corr/MarcheggianiFT17}, which already achieves state-of-the-art results on the CoNLL-2009 English dataset.
Following their approach we employ the same bidirectional (BiLSTM) encoder and enrich it with a syntactic GCN.

The CoNLL-2009 benchmark assumes that predicate positions are already marked in the test set (e.g., we would know that {\it makes}, {\it repairs} and {\it engines} in Figure~1 are predicates), so no predicate identification is needed.
Also, as we focus here solely  on identifying arguments and labeling them with semantic roles, for predicate disambiguation (i.e., marking {\it makes} as {\it make.01}) we use of an off-the-shelf disambiguation model \cite{roth-lapata:2016:P16-1,DBLP:conf/conll/BjorkelundHN09}.  As in \newcite{DBLP:journals/corr/MarcheggianiFT17} and in most previous work,
we process individual predicates in isolation, so for each predicate, our task reduces to a sequence labeling problem. That is, given a predicate (e.g., {\it disputed} in Figure~\ref{fig:model}) one needs to identify and label all its arguments (e.g., label {\it estimates} as A1 and label {\it those} as `NULL', indicating that {\it those} is not an argument of {\it disputed}).

The semantic role labeler we propose is composed of four components (see Figure \ref{fig:model}):
\begin{itemize}
\item  look-ups of word embeddings;
\item  a BiLSTM encoder that takes  as input the word representation of each word in a sentence;
\item a syntax-based GCN encoder that re-encodes the BiLSTM representation based on the automatically predicted syntactic structure of the sentence;
\item  a role classifier that takes as input the GCN representation of the candidate argument and the representation of the predicate to predict the role associated with the candidate word.
\end{itemize}

\subsection{Word representations} \label{sec:word-repr}
For each word $w_i$ in the considered sentence,  we create a sentence-specific word representation $x_i$.
We represent each word $w$ as the concatenation of four vectors:\footnote{We drop the index $i$ from the notation for the sake of brevity.} a randomly initialized word embedding $x^{re} \in \mathbb{R}^{d_w}$, a pre-trained word embedding $x^{pe} \in \mathbb{R}^{d_w}$ estimated on an external text collection, a randomly initialized  part-of-speech tag embedding $x^{pos} \in \mathbb{R}^{d_p}$ and a randomly initialized lemma embedding $x^{le} \in \mathbb{R}^{d_l}$ (active only if the word is a predicate).
The randomly initialized embeddings $x^{re}$, $x^{pos}$, and $x^{le}$ are fine-tuned during training, while the pre-trained ones are kept fixed.
The final word representation is given by $x = x^{re} \circ x^{pe} \circ x^{pos} \circ x^{le}$, where $\circ$ represents the concatenation operator.

\subsection{Bidirectional LSTM layer}
One of the most popular and effective ways to represent sequences, such as sentences \cite{DBLP:conf/interspeech/MikolovKBCK10}, is to use recurrent neural networks (RNN) \cite{DBLP:journals/cogsci/Elman90}.
In particular their gated versions, Long Short-Term Memory (LSTM) networks  \cite{DBLP:journals/neco/HochreiterS97} and Gated Recurrent Units (GRU) \cite{DBLP:conf/emnlp/ChoMGBBSB14}, have proven  effective in modeling long sequences \cite{DBLP:journals/tacl/ChiuN16,DBLP:conf/nips/SutskeverVL14}. 

Formally, an LSTM can be defined as a function $LSTM_{\theta}(x_{1:i})$ that takes as input the sequence $x_{1:i}$ and returns a hidden state $h_i \in \mathbb{R}^{d_h}$. 
This state can be regarded as a representation of the sentence from the start to the position $i$, or, in other words, it encodes the word at position $i$ along with its left context. 
However, the right context is also important, so  
Bidirectional LSTMs \cite{graves2008supervised} use two LSTMs: one for the forward pass, and another for the backward pass, $LSTM_{F}$ and $LSTM_{B}$, respectively. 
By concatenating the states of both LSTMs, we create a complete context-aware representation of a word $BiLSTM(x_{1:n},i) = LSTM_{F}(x_{1:i}) \circ LSTM_{B}(x_{n:i})$. We follow \newcite{DBLP:journals/corr/MarcheggianiFT17} and stack $J$ layers of bidirectional LSTMs, where each layer takes the lower layer as its input. 

\subsection{Graph convolutional layer}
The representation calculated with the BiLSTM encoder is fed as input to a GCN of the form defined in Equation (\ref{eq:final-gcn}).
The neighboring nodes of a node $v$, namely $\mathcal{N}(v)$, and their relations to $v$ are predicted by an external syntactic parser.

\subsection{Semantic role classifier}
The classifier predicts semantic roles of words given the predicate while relying on word representations provided by GCN; we concatenate hidden states of the candidate argument word and the predicate word and use them as input to a classifier (Figure \ref{fig:model}, top). 
The softmax classifier computes the probability of the role (including special `NULL' role):\begin{equation} \label{eq:softmax}
p(r|t_i, t_p, l) \propto \exp(W_{l,r} ( t_i \circ t_p)),
\end{equation}
where $t_i$ and $t_p$ are representations produced by the graph convolutional encoder, $l$ is the lemma of predicate $p$, and the symbol $\propto$ signifies proportionality.\footnote{We abuse the notation and refer as $p$ both to the predicate word and to its position in the sentence.}
As \newcite{fitzgerald-EtAl:2015:EMNLP} and \newcite{DBLP:journals/corr/MarcheggianiFT17}, instead of using a fixed matrix $W_{l,r}$ or simply assuming that $W_{l,r} = W_r$,
we jointly embed the role $r$ and predicate lemma $l$ using a non-linear transformation:
\begin{equation}
\label{eq:frame_role}
 W_{l,r} =  ReLU(U ({q}_l\circ {q}_r)),
 \end{equation}
where $U$ is a parameter matrix, whereas ${q}_l\in \mathbb{R}^{d'_l}$ and ${q}_r \in \mathbb{R}^{d_r}$ are randomly initialized embeddings of predicate lemmas and roles.
In this way each role prediction is predicate-specific, and, at the same time, we expect to learn a good representation for roles associated with infrequent predicates.  As our training objective we use the categorical cross-entropy.

\section{Experiments}
\label{sec:experiments}

\subsection{Datasets and parameters}
\label{subsect:datasets}

\begin{figure}
\begin{center}
\includegraphics[width=\columnwidth]{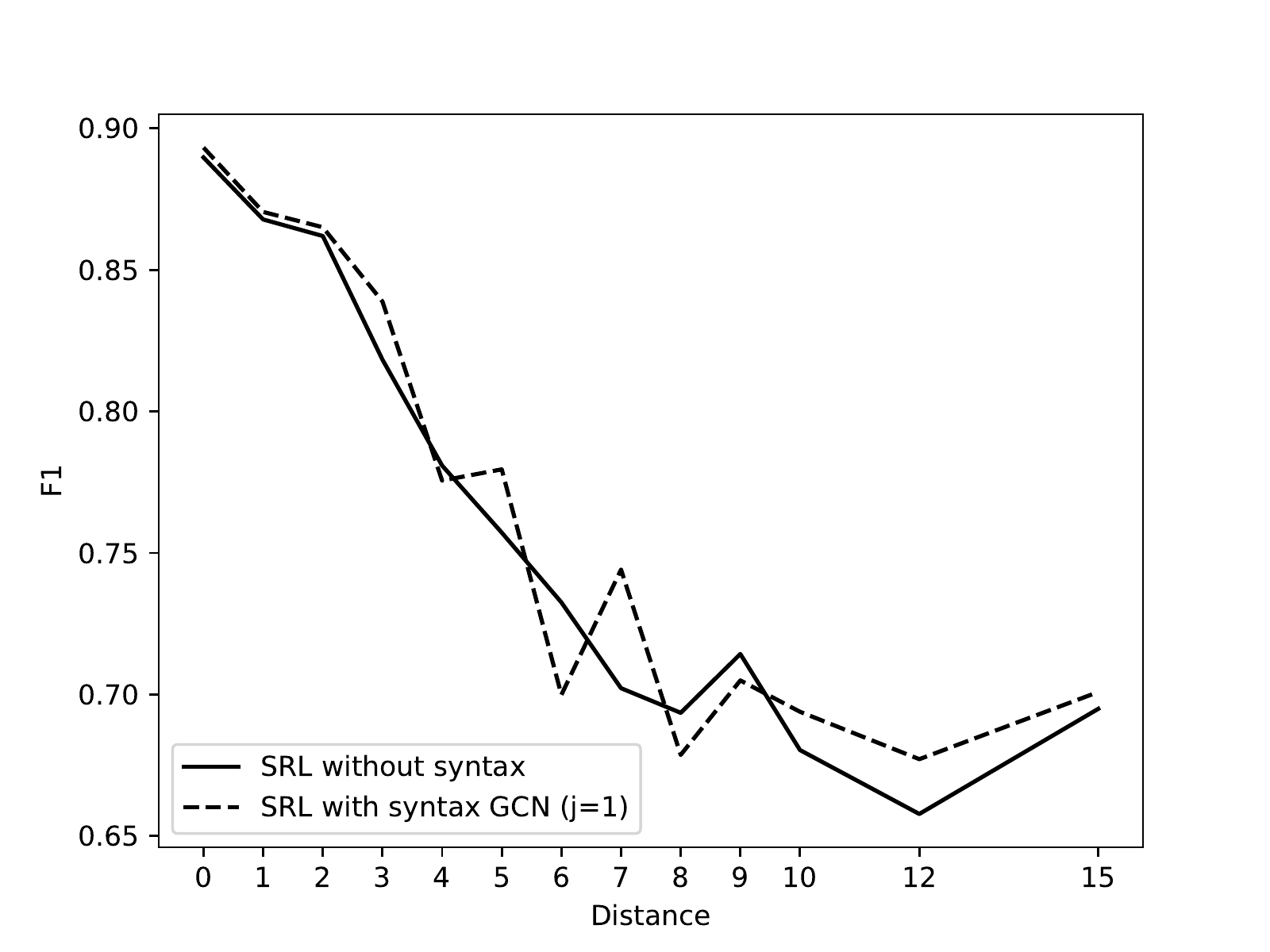}
\caption{\label{fig:token_dist} F\textsubscript{1} as function of word distance. The distance starts from zero, since nominal predicates can be arguments of themselves.
} 
\end{center}
\end{figure}

We tested the proposed SRL model on the English and Chinese
CoNLL-2009 dataset with standard splits into training, test and development sets. 
The predicted POS tags for both languages were provided by the CoNLL-2009 shared-task organizers.
For the predicate disambiguator we used the ones from \newcite{roth-lapata:2016:P16-1} for English and from \newcite{DBLP:conf/conll/BjorkelundHN09} for Chinese.
We parsed English sentences with the BIST Parser \cite{TACL885}, whereas for Chinese we used  automatically predicted parses provided by the CoNLL-2009 shared-task organizers.

For English, we used external embeddings of \newcite{DBLP:conf/acl/DyerBLMS15}, learned using the structured skip n-gram approach of \newcite{ling-EtAl:2015:NAACL-HLT}. 
For Chinese we used external embeddings produced with the neural language model of \newcite{DBLP:journals/jmlr/BengioDVJ03}.
We used  \textit{edge dropout} in GCN: when computing $h^{(k)}_v$, we ignore
each node $v \in \mathcal{N}(v)$ with probability $\beta$.
Adam \cite{kingma2014adam} was used as an optimizer.
The hyperparameter tuning and all model selection were performed on the English development set; the chosen values are shown in Appendix.

\subsection{Results and discussion}

\begin{figure}
\begin{center}
\includegraphics[width=\columnwidth]{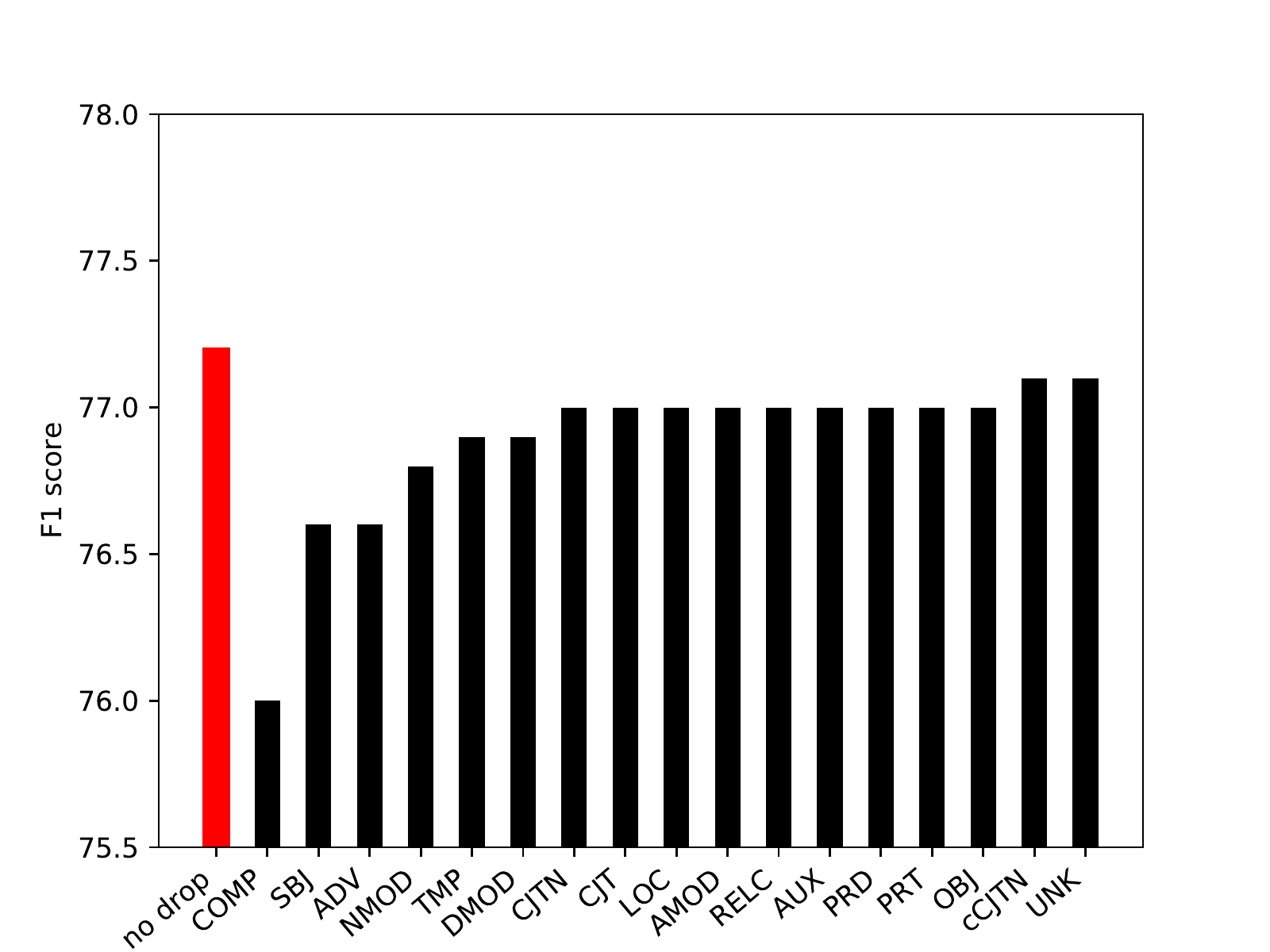}
\caption{\label{fig:dep-label-ch} Performance with dependency arcs of given type dropped, on Chinese development set.} 
\end{center}
\end{figure}

\begin{table}[t]
\centering
\scalebox{0.86}{
\begin{tabular}{@{\extracolsep{\fill}}l@{\hspace{6pt}}l@{\hspace{6pt}}c@{\hspace{6pt}}c@{\hspace{6pt}}c}
  \toprule
   & System (English) & P & R & {F\textsubscript{1}}   \\
   \midrule
   & LSTMs  & 84.3 & 81.1 &  82.7 \\
   & LSTMs + GCNs (K=1) & {85.2} & {81.6} & {83.3} \\
   & LSTMs + GCNs (K=2) & {84.1} & {81.4} & {82.7} \\
   & LSTMs + GCNs (K=1), no gates & {84.7} & {81.4} & {83.0} \\
   \midrule
   & GCNs (no LSTMs), K=1 & 79.9 & 70.4 & 74.9 \\
   & GCNs (no LSTMs), K=2 & 83.4 & 74.6 & 78.7 \\
   & GCNs (no LSTMs), K=3  & 83.6 & 75.8 & 79.5 \\
   & GCNs (no LSTMs), K=4  & 82.7 & 76.0 & 79.2 \\
\bottomrule
\end{tabular}
}
\caption{\label{tab:dev-results} SRL results without predicate disambiguation on the English development set. 
}
\end{table}

\begin{table}[t]
\centering
\scalebox{0.86}{
\begin{tabular}{@{\extracolsep{\fill}}l@{\hspace{6pt}}l@{\hspace{6pt}}c@{\hspace{6pt}}c@{\hspace{6pt}}c}
  \toprule
   & System (Chinese) & P & R & {F\textsubscript{1}}   \\
   \midrule
   & LSTMs & 78.3 & 72.3 &  75.2 \\
   & LSTMs + GCNs (K=1) & {79.9} & {74.4} & {77.1} \\
   & LSTMs + GCNs (K=2) & {78.7} & {74.0} & {76.2} \\
   & LSTMs + GCNs (K=1), no gates & {78.2} & {74.8} & {76.5} \\
   \midrule
   & GCNs (no LSTMs), K=1 & 78.7 & 58.5 & 67.1 \\
   & GCNs (no LSTMs), K=2 & 79.7 & 62.7 & 70.1 \\
   & GCNs (no LSTMs), K=3  & 76.8 & 66.8 & 71.4 \\
   & GCNs (no LSTMs), K=4  & 79.1 & 63.5 & 70.4 \\
\bottomrule
\end{tabular}
}
\caption{\label{tab:dev-results-ch} SRL results without predicate disambiguation on the Chinese development set. 
}
\end{table}

\begin{table}[t]
\centering
\scalebox{0.86}{
\begin{tabular}{@{\extracolsep{\fill}}l@{\hspace{6pt}}l@{\hspace{6pt}}c@{\hspace{6pt}}c@{\hspace{6pt}}c}
  \toprule
   & System & P & R & {F\textsubscript{1}}   \\
   \midrule
   &\newcite{lei-EtAl:2015:NAACL-HLT} \small{(local)} & - & - & 86.6 \\
   &\newcite{fitzgerald-EtAl:2015:EMNLP} \small{(local)} & - & - & 86.7 \\
   & \newcite{roth-lapata:2016:P16-1} \small{(local)} & 88.1 & 85.3 & 86.7 \\
   & \newcite{DBLP:journals/corr/MarcheggianiFT17} \small (local) & 88.6 & 86.7 &  87.6 \\
   & {\bf Ours} \small{\bf (local)} & {\bf 89.1} & {\bf 86.8} & {\bf 88.0} \\
   \midrule
   &\newcite{bjorkelund2010high} \small{(global)} & 88.6 & 85.2 & 86.9 \\
      &\newcite{fitzgerald-EtAl:2015:EMNLP} \small{(global)} & - & - & 87.3 \\
      &\newcite{foland2015} \small{(global)} & - & - & 86.0 \\
  &\newcite{DBLP:conf/conll/SwayamdiptaBDS16} \small{(global)} & - & - & 85.0 \\
   &\newcite{roth-lapata:2016:P16-1} \small{(global)} & 90.0 & 85.5 & 87.7 \\
   \midrule
   &\newcite{fitzgerald-EtAl:2015:EMNLP} \small{(ensemble)} & - & - & 87.7 \\
   &\newcite{roth-lapata:2016:P16-1} \small{(ensemble)} & 90.3 & 85.7 & 87.9 \\
   & {\bf Ours} \small{\bf (ensemble 3x)} & {\bf 90.5} & {\bf 87.7} & {\bf 89.1} \\
\bottomrule
\end{tabular}
}
\caption{\label{tab:wsj-results} Results on the test set  for English. 
}
\end{table}

\begin{table}[t]
\centering
\scalebox{0.86}{
\begin{tabular}{@{\extracolsep{\fill}}l@{\hspace{6pt}}l@{\hspace{6pt}}c@{\hspace{6pt}}c@{\hspace{6pt}}c}
  \toprule
   & System & P & R & {F\textsubscript{1}}   \\
   \midrule
   &\newcite{DBLP:conf/conll/ZhaoCKUT09} (global) & 80.4 & 75.2 & 77.7 \\
   &\newcite{DBLP:conf/conll/BjorkelundHN09} (global)  & 82.4 & 75.1 & 78.6 \\
   & \newcite{roth-lapata:2016:P16-1} (global)  & 83.2 & 75.9 & 79.4 \\
   & {\bf Ours} \small{\bf (local)}& {\bf 84.6} & {\bf 80.4} & {\bf 82.5} \\
\bottomrule
\end{tabular}
}
\caption{\label{tab:chinese-results} Results on the Chinese test set. 
}
\end{table}

In order to show that GCN layers are effective, we first compare our model against its version which lacks GCN layers (i.e. essentially the model of \newcite{DBLP:journals/corr/MarcheggianiFT17}). Importantly, to measure the genuine contribution of GCNs, we first tuned this syntax-agnostic model (e.g., the number of LSTM layers) to get best possible performance on the development set.\footnote{For example, if we would have used only one layer of LSTMs, gains from using GCNs would be even larger.}

We compare  the syntax-agnostic model with 3 syntax-aware versions: one GCN layer over syntax ($K=1$), one layer GCN without gates and  two GCN layers ($K=2$).
As we rely on the same off-the-shelf disambiguator for all versions of the model,  in Table \ref{tab:dev-results} and \ref{tab:dev-results-ch} we report SRL-only scores (i.e., predicate disambiguation is not evaluated) on the English and Chinese development sets.
For both datasets, the syntax-aware model with one GCN layers ($K=1$) performs the best, outperforming the LSTM version by 1.9\% and 0.6\% for Chinese and English, respectively.
The reasons why the improvements on Chinese are much larger are not entirely clear (e.g., both languages are relative fixed word order ones, and the syntactic parses for Chinese are considerably less accurate),
 this may be attributed to a higher proportion of long-distance dependencies between predicates and arguments in Chinese (see Section~\ref{sec:complement}).
Edge-wise gating (Section \ref{sec:gating}) also appears important:
removing gates leads to a drop of 0.3\% {F\textsubscript{1}} for English and 0.6\% {F\textsubscript{1}} for Chinese.

Stacking two GCN layers does not give any benefit.
When BiLSTM layers are dropped altogether, stacking two layers ($K=2$) of GCNs greatly improves the performance, resulting in a 3.8\% jump in {F\textsubscript{1}} for English and a 3.0\% jump in {F\textsubscript{1}} for Chinese.
Adding a 3rd layer of GCN ($K=3$) further improves the performance.\footnote{Note that GCN layers are computationally cheaper than LSTM ones, even in our non-optimized implementation.} 
This suggests that extra GCN layers are effective but largely redundant with respect to what LSTMs already capture.

In Figure \ref{fig:token_dist}, we show the $F_1$ scores results on the English development set as a function of the distance, in terms of tokens, between a candidate argument and its predicate.
As expected, GCNs appear to be more beneficial for long distance dependencies, as shorter ones are already accurately captured by the LSTM encoder. 

We looked closer in contribution of specific dependency relations for Chinese.
In order to assess this without retraining the model multiple times, we drop all dependencies of a given type at test time (one type at a time, only for types appearing over 300 times in the development set) and observe changes in
performance.
In Figure \ref{fig:dep-label-ch}, we see that the most informative dependency is COMP (complement).
Relative clauses in Chinese are very frequent and typically marked with particle \begin{CJK*}{UTF8}{gbsn}的\end{CJK*} (de). The relative clause will syntactically depend on \begin{CJK*}{UTF8}{gbsn}的\end{CJK*} as COMP, so COMP encodes important information about predicate-argument structure. These are often long-distance dependencies and may not be accurately captured by LSTMs.
Although TMP (temporal) dependencies are not as frequent ($\sim$2\% of all dependencies), they are also important:  temporal information is mirrored in semantic roles.

In order to compare to previous work, in Table~\ref{tab:wsj-results} we report test results on the English in-domain (WSJ) evaluation data.
Our model is {\it local}, as all the argument detection and labeling decisions are conditionally independent: their interaction is captured solely by the LSTM+GCN encoder. This makes our model fast and simple, though, as shown in previous work, {\it global} modeling of the structured output is beneficial.\footnote{As seen in
Table \ref{tab:wsj-results}, labelers of \newcite{fitzgerald-EtAl:2015:EMNLP} and \newcite{roth-lapata:2016:P16-1} gained 0.6-1.0\%.} We leave this extension for future work. Interestingly, we outperform even the best global model and the best ensemble of global models, without using global modeling or ensembles. When we create an ensemble of 3 models with the product-of-expert combination rule, we improve by 1.2\% over the best previous result, achieving 89.1\% F\textsubscript{1}.\footnote{To compare to previous work, we report combined scores which also include predicate disambiguation. As we use
disambiguators from previous work (see Section~\ref{subsect:datasets}), actual gains in argument identification and labeling are even larger.}

For Chinese (Table \ref{tab:chinese-results}), our best model outperforms the state-of-the-art model of  \newcite{roth-lapata:2016:P16-1} by even larger margin of 3.1\%.

For English, in the CoNLL shared task, 
systems are also evaluated on the out-of-domain dataset.
Statistical models are typically less accurate when they are applied to out-of-domain data.
Consequently, the predicted syntax for the out-of-domain test set is of lower quality, which  negatively affects the quality of GCN embeddings.
However, our model works surprisingly well on out-of-domain data (Table \ref{tab:ood-results}), substantially outperforming all the previous syntax-aware models. 
This suggests that our model is fairly robust to mistakes in syntax. 
As expected though, our model does not outperform the syntax-agnostic model of \newcite{DBLP:journals/corr/MarcheggianiFT17}.

\begin{table}[t]
\centering
\scalebox{0.86}{
\begin{tabular}{@{\extracolsep{\fill}}l@{\hspace{6pt}}l@{\hspace{6pt}}c@{\hspace{6pt}}c@{\hspace{6pt}}c}
  \toprule
   & System & P & R & {F\textsubscript{1}}   \\
   \midrule
   &\newcite{lei-EtAl:2015:NAACL-HLT} \small{(local)} & - & - & 75.6 \\
   &\newcite{fitzgerald-EtAl:2015:EMNLP} \small{(local)} & - & - & 75.2 \\
   & \newcite{roth-lapata:2016:P16-1} \small{(local)} & 76.9 & 73.8 & 75.3 \\
   & \newcite{DBLP:journals/corr/MarcheggianiFT17} \small (local) & 78.9 &  75.7 &  77.3 \\
   & {\bf Ours} \small{\bf (local)} & {\bf 78.5} & {\bf 75.9} & {\bf 77.2} \\
   \midrule
   &\newcite{bjorkelund2010high} \small{(global)} & 77.9 & 73.6 & 75.7 \\
   &\newcite{fitzgerald-EtAl:2015:EMNLP} \small{(global)} & - & - & 75.2 \\
         &\newcite{foland2015} \small{(global)} & - & - & 75.9 \\
   &\newcite{roth-lapata:2016:P16-1} \small{(global)} & 78.6 & 73.8 & 76.1 \\
   \midrule
   &\newcite{fitzgerald-EtAl:2015:EMNLP} \small{(ensemble)} & - & - & 75.5 \\
   &\newcite{roth-lapata:2016:P16-1} \small{(ensemble)} & 79.7 & 73.6 & 76.5 \\
   & {\bf Ours} \small{\bf (ensemble 3x)} & {\bf 80.8} & {\bf 77.1} & {\bf 78.9} \\
\bottomrule
\end{tabular}
}
\caption{\label{tab:ood-results} Results on the out-of-domain test set. 
}
\end{table}

\section{Related Work}\label{sec:related-work}

\noindent
Perhaps the earliest methods modeling syntax-semantics interface with RNNs are due to~\cite{HendersonConll08,TitovIjcai09,TitovCoNLL09ST}, they used shift-reduce parsers for joint SRL and syntactic parsing, and relied on RNNs to model statistical dependencies across syntactic and semantic parsing actions. A more modern (e.g., based on LSTMs) and effective reincarnation of this line of research has been proposed in ~\newcite{DBLP:conf/conll/SwayamdiptaBDS16}.
Other recent work which considered incorporation of syntactic information in neural SRL models include: \newcite{fitzgerald-EtAl:2015:EMNLP} who use standard syntactic features within an MLP calculating potentials of a CRF model;
\newcite{roth-lapata:2016:P16-1} who enriched standard features for SRL with  LSTM representations of syntactic paths between arguments and predicates;
\newcite{lei-EtAl:2015:NAACL-HLT} who relied on low-rank tensor factorizations for modeling syntax. Also \newcite{foland2015} used  (non-graph) convolutional  networks and provided syntactic features 
as input. 
A very different line of research, but with similar goals to ours (i.e. integrating syntax with minimal feature engineering), used tree kernels~\cite{DBLP:journals/coling/MoschittiPB08}. 

Beyond SRL, there have been many proposals on how to incorporate syntactic information in RNN models, for example, in the context of neural machine translation~\cite{eriguchi2017learning,DBLP:conf/wmt/SennrichH16}. 
One of the most popular and attractive approaches is to use tree-structured recursive neural networks~\cite{socher-EtAl:2013:EMNLP,DBLP:conf/emnlp/LeZ14,DBLP:conf/acl/DyerBLMS15},
including stacking them on top of a sequential BiLSTM \cite{DBLP:conf/acl/MiwaB16}.
An approach of \newcite{DBLP:conf/emnlp/MouPLXZJ15} to sentiment analysis and question classification,
introduced even before GCNs became popular in the machine learning community, is related to graph convolution. However, 
it is inherently single-layer and tree-specific, uses bottom-up computations, does not share parameters across syntactic functions and does not use gates. Gates have been previously used in GCNs~\cite{li2015gated} but between GCN layers rather than for individual edges.

Previous approaches to integrating syntactic information in neural models are mainly designed to induce representations of sentences or syntactic constituents. In contrast, the approach we presented incorporates syntactic information at word level. This may be attractive from the engineering perspective, as it can be used, as we have shown, instead or along with RNN models.

\section{Conclusions and Future Work}\label{sec:conclusion}

We demonstrated how GCNs can be used to incorporate syntactic information in neural models and specifically to construct a syntax-aware SRL model, resulting in state-of-the-art results for Chinese and English. 
There are relatively straightforward steps which can further improve the SRL results. For example, we relied on labeling arguments independently, whereas using a joint model is likely to significantly improve the performance. 

More generally, given  simplicity of GCNs
and their applicability to general graph structures (not necessarily trees), we believe that there are many NLP tasks where GCNs  can be used to incorporate linguistic structures  (e.g., syntactic and semantic representations of sentences and discourse parses or co-reference graphs for documents). 

 \subsection*{Acknowledgements}
 We would thank Anton Frolov, Michael Schlichtkrull, Thomas Kipf, Michael Roth, Max Welling, Yi Zhang, and Wilker Aziz for their suggestions and comments.
 The project was supported by the European Research Council (ERC StG BroadSem 678254), the Dutch National Science Foundation (NWO VIDI 639.022.518) and an Amazon Web Services (AWS) grant.

\bibliography{biblio}

\begin{thebibliography}{}
\expandafter\ifx\csname natexlab\endcsname\relax\def\natexlab#1{#1}\fi

\bibitem[{Bengio et~al.(2003)Bengio, Ducharme, Vincent, and
  Janvin}]{DBLP:journals/jmlr/BengioDVJ03}
Yoshua Bengio, R{\'{e}}jean Ducharme, Pascal Vincent, and Christian Janvin.
  2003.
\newblock A neural probabilistic language model.
\newblock {\em Journal of Machine Learning Research\/} 3:1137--1155.

\bibitem[{Bj{\"o}rkelund et~al.(2010)Bj{\"o}rkelund, Bohnet, Hafdell, and
  Nugues}]{bjorkelund2010high}
Anders Bj{\"o}rkelund, Bernd Bohnet, Love Hafdell, and Pierre Nugues. 2010.
\newblock A high-performance syntactic and semantic dependency parser.
\newblock In {\em Proceedings of COLING: Demonstrations\/}.

\bibitem[{Bj{\"{o}}rkelund et~al.(2009)Bj{\"{o}}rkelund, Hafdell, and
  Nugues}]{DBLP:conf/conll/BjorkelundHN09}
Anders Bj{\"{o}}rkelund, Love Hafdell, and Pierre Nugues. 2009.
\newblock Multilingual semantic role labeling.
\newblock In {\em Proceedings of CoNLL\/}.

\bibitem[{Chiu and Nichols(2016)}]{DBLP:journals/tacl/ChiuN16}
Jason P.~C. Chiu and Eric Nichols. 2016.
\newblock Named entity recognition with bidirectional {LSTM-CNN}s.
\newblock {\em {TACL}\/} 4:357--370.

\bibitem[{Cho et~al.(2014)Cho, van Merrienboer, G{\"{u}}l{\c{c}}ehre, Bahdanau,
  Bougares, Schwenk, and Bengio}]{DBLP:conf/emnlp/ChoMGBBSB14}
Kyunghyun Cho, Bart van Merrienboer, {\c{C}}aglar G{\"{u}}l{\c{c}}ehre, Dzmitry
  Bahdanau, Fethi Bougares, Holger Schwenk, and Yoshua Bengio. 2014.
\newblock Learning phrase representations using {RNN} encoder-decoder for
  statistical machine translation.
\newblock In {\em Proceedings of EMNLP\/}.

\bibitem[{Dauphin et~al.(2016)Dauphin, Fan, Auli, and
  Grangier}]{DBLP:journals/corr/DauphinFAG16}
Yann~N. Dauphin, Angela Fan, Michael Auli, and David Grangier. 2016.
\newblock Language modeling with gated convolutional networks.
\newblock {\em arXiv preprint arXiv:1612.08083\/} .

\bibitem[{Duvenaud et~al.(2015)Duvenaud, Maclaurin, Aguilera{-}Iparraguirre,
  Bombarell, Hirzel, Aspuru{-}Guzik, and Adams}]{NIPS2015_5954}
David~K. Duvenaud, Dougal Maclaurin, Jorge Aguilera{-}Iparraguirre, Rafael
  Bombarell, Timothy Hirzel, Al{\'{a}}n Aspuru{-}Guzik, and Ryan~P. Adams.
  2015.
\newblock Convolutional networks on graphs for learning molecular fingerprints.
\newblock In {\em Proceedings of NIPS\/}.

\bibitem[{Dyer et~al.(2015)Dyer, Ballesteros, Ling, Matthews, and
  Smith}]{DBLP:conf/acl/DyerBLMS15}
Chris Dyer, Miguel Ballesteros, Wang Ling, Austin Matthews, and Noah~A. Smith.
  2015.
\newblock Transition-based dependency parsing with stack long short-term
  memory.
\newblock In {\em Proceedings of ACL\/}.

\bibitem[{Elman(1990)}]{DBLP:journals/cogsci/Elman90}
Jeffrey~L. Elman. 1990.
\newblock Finding structure in time.
\newblock {\em Cognitive Science\/} 14(2):179--211.

\bibitem[{Eriguchi et~al.(2017)Eriguchi, Tsuruoka, and
  Cho}]{eriguchi2017learning}
Akiko Eriguchi, Yoshimasa Tsuruoka, and Kyunghyun Cho. 2017.
\newblock Learning to parse and translate improves neural machine translation.
\newblock {\em arXiv preprint arXiv:1702.03525\/} .

\bibitem[{FitzGerald et~al.(2015)FitzGerald, T\"{a}ckstr\"{o}m, Ganchev, and
  Das}]{fitzgerald-EtAl:2015:EMNLP}
Nicholas FitzGerald, Oscar T\"{a}ckstr\"{o}m, Kuzman Ganchev, and Dipanjan Das.
  2015.
\newblock Semantic role labeling with neural network factors.
\newblock In {\em Proceedings of EMNLP\/}.

\bibitem[{Foland and Martin(2015)}]{foland2015}
William Foland and James Martin. 2015.
\newblock Dependency-based semantic role labeling using convolutional neural
  networks.
\newblock In {\em Proceedings of the Fourth Joint Conference on Lexical and
  Computational Semantics {(*SEM)}\/}.

\bibitem[{Gesmundo et~al.(2009)Gesmundo, Henderson, Merlo, and
  Titov}]{TitovCoNLL09ST}
Andrea Gesmundo, James Henderson, Paola Merlo, and Ivan Titov. 2009.
\newblock Latent variable model of synchronous syntactic-semantic parsing for
  multiple languages.
\newblock In {\em Proceedings of CoNLL\/}.

\bibitem[{Gildea and Jurafsky(2002)}]{gildea2002automatic}
Daniel Gildea and Daniel Jurafsky. 2002.
\newblock Automatic labeling of semantic roles.
\newblock {\em Computational linguistics\/} 28(3):245--288.

\bibitem[{Gilmer et~al.(2017)Gilmer, Schoenholz, Riley, Vinyals, and
  Dahl}]{gilmer2017neural}
Justin Gilmer, Samuel~S. Schoenholz, Patrick~F. Riley, Oriol Vinyals, and
  George~E. Dahl. 2017.
\newblock Neural message passing for quantum chemistry.
\newblock {\em arXiv preprint arXiv:1704.01212\/} .

\bibitem[{Graves(2008)}]{graves2008supervised}
Alex Graves. 2008.
\newblock {\em Supervised sequence labelling with recurrent neural networks\/}.
\newblock Ph.D. thesis, M{\"u}nchen, Techn. Univ., Diss., 2008.

\bibitem[{Henderson et~al.(2008)Henderson, Merlo, Musillo, and
  Titov}]{HendersonConll08}
James Henderson, Paola Merlo, Gabriele Musillo, and Ivan Titov. 2008.
\newblock A latent variable model of synchronous parsing for syntactic and
  semantic dependencies.
\newblock In {\em Proceedings of CoNLL\/}.

\bibitem[{Hochreiter and Schmidhuber(1997)}]{DBLP:journals/neco/HochreiterS97}
Sepp Hochreiter and J{\"{u}}rgen Schmidhuber. 1997.
\newblock Long short-term memory.
\newblock {\em Neural Computation\/} 9(8):1735--1780.

\bibitem[{Johansson and Nugues(2008)}]{DBLP:conf/coling/JohanssonN08}
Richard Johansson and Pierre Nugues. 2008.
\newblock The effect of syntactic representation on semantic role labeling.
\newblock In {\em Proceedings of {COLING}\/}.

\bibitem[{Kearnes et~al.(2016)Kearnes, McCloskey, Berndl, Pande, and
  Riley}]{kearnes2016molecular}
Steven Kearnes, Kevin McCloskey, Marc Berndl, Vijay Pande, and Patrick Riley.
  2016.
\newblock Molecular graph convolutions: moving beyond fingerprints.
\newblock {\em Journal of computer-aided molecular design\/} 30(8):595--608.

\bibitem[{Kingma and Ba(2015)}]{kingma2014adam}
Diederik Kingma and Jimmy Ba. 2015.
\newblock Adam: A method for stochastic optimization.
\newblock In {\em Proceedings of ICLR\/}.

\bibitem[{Kiperwasser and Goldberg(2016)}]{TACL885}
Eliyahu Kiperwasser and Yoav Goldberg. 2016.
\newblock Simple and accurate dependency parsing using bidirectional {LSTM}
  feature representations.
\newblock {\em TACL\/} 4:313--327.

\bibitem[{Kipf and Welling(2017)}]{DBLP:journals/corr/KipfW16}
Thomas~N. Kipf and Max Welling. 2017.
\newblock Semi-supervised classification with graph convolutional networks.
\newblock In {\em Proceedings of ICLR\/}.

\bibitem[{Le and Zuidema(2014)}]{DBLP:conf/emnlp/LeZ14}
Phong Le and Willem Zuidema. 2014.
\newblock The inside-outside recursive neural network model for dependency
  parsing.
\newblock In {\em Proceedings of EMNLP\/}.

\bibitem[{LeCun et~al.(2001)LeCun, Bottou, Bengio, and Haffner}]{lecun-01a}
Yann LeCun, Leon Bottou, Yoshua Bengio, and Patrick Haffner. 2001.
\newblock Gradient-based learning applied to document recognition.
\newblock In {\em Proceedings of Intelligent Signal Processing\/}.

\bibitem[{Lei et~al.(2015)Lei, Zhang, M\`{a}rquez, Moschitti, and
  Barzilay}]{lei-EtAl:2015:NAACL-HLT}
Tao Lei, Yuan Zhang, Llu\'{i}s M\`{a}rquez, Alessandro Moschitti, and Regina
  Barzilay. 2015.
\newblock High-order low-rank tensors for semantic role labeling.
\newblock In {\em Proceedings of NAACL\/}.

\bibitem[{Levin(1993)}]{levin1993english}
Beth Levin. 1993.
\newblock {\em English verb classes and alternations: A preliminary
  investigation\/}.
\newblock University of Chicago press.

\bibitem[{Li et~al.(2016)Li, Tarlow, Brockschmidt, and Zemel}]{li2015gated}
Yujia Li, Daniel Tarlow, Marc Brockschmidt, and Richard~S. Zemel. 2016.
\newblock Gated graph sequence neural networks.
\newblock In {\em Proceedings of ICLR\/}.

\bibitem[{Ling et~al.(2015)Ling, Dyer, Black, and
  Trancoso}]{ling-EtAl:2015:NAACL-HLT}
Wang Ling, Chris Dyer, Alan~W Black, and Isabel Trancoso. 2015.
\newblock Two/too simple adaptations of word2vec for syntax problems.
\newblock In {\em Proceedings of NAACL\/}.

\bibitem[{Linzen et~al.(2016)Linzen, Dupoux, and
  Goldberg}]{DBLP:journals/tacl/LinzenDG16}
Tal Linzen, Emmanuel Dupoux, and Yoav Goldberg. 2016.
\newblock Assessing the ability of {LSTMs} to learn syntax-sensitive
  dependencies.
\newblock {\em {TACL}\/} 4:521--535.

\bibitem[{Marcheggiani et~al.(2017)Marcheggiani, Frolov, and
  Titov}]{DBLP:journals/corr/MarcheggianiFT17}
Diego Marcheggiani, Anton Frolov, and Ivan Titov. 2017.
\newblock A simple and accurate syntax-agnostic neural model for
  dependency-based semantic role labeling.
\newblock {\em arXiv preprint arXiv:1701.02593\/} .

\bibitem[{Mikolov et~al.(2010)Mikolov, Karafi{\'{a}}t, Burget, Cernock{\'{y}},
  and Khudanpur}]{DBLP:conf/interspeech/MikolovKBCK10}
Tomas Mikolov, Martin Karafi{\'{a}}t, Luk{\'{a}}s Burget, Jan Cernock{\'{y}},
  and Sanjeev Khudanpur. 2010.
\newblock Recurrent neural network based language model.
\newblock In {\em Proceedings of INTERSPEECH\/}.

\bibitem[{Miwa and Bansal(2016)}]{DBLP:conf/acl/MiwaB16}
Makoto Miwa and Mohit Bansal. 2016.
\newblock End-to-end relation extraction using lstms on sequences and tree
  structures.
\newblock In {\em Proceedings of ACL\/}.

\bibitem[{Moschitti et~al.(2008)Moschitti, Pighin, and
  Basili}]{DBLP:journals/coling/MoschittiPB08}
Alessandro Moschitti, Daniele Pighin, and Roberto Basili. 2008.
\newblock Tree kernels for semantic role labeling.
\newblock {\em Computational Linguistics\/} 34(2):193--224.

\bibitem[{Mou et~al.(2015)Mou, Peng, Li, Xu, Zhang, and
  Jin}]{DBLP:conf/emnlp/MouPLXZJ15}
Lili Mou, Hao Peng, Ge~Li, Yan Xu, Lu~Zhang, and Zhi Jin. 2015.
\newblock Discriminative neural sentence modeling by tree-based convolution.
\newblock In {\em Proceedings of EMNLP\/}.

\bibitem[{Pradhan et~al.(2005)Pradhan, Hacioglu, Ward, Martin, and
  Jurafsky}]{DBLP:conf/conll/PradhanHWMJ05}
Sameer Pradhan, Kadri Hacioglu, Wayne~H. Ward, James~H. Martin, and Daniel
  Jurafsky. 2005.
\newblock Semantic role chunking combining complementary syntactic views.
\newblock In {\em Proceedings of CoNLL\/}.

\bibitem[{Punyakanok et~al.(2008)Punyakanok, Roth, and
  Yih}]{DBLP:journals/coling/PunyakanokRY08}
Vasin Punyakanok, Dan Roth, and Wen{-}tau Yih. 2008.
\newblock The importance of syntactic parsing and inference in semantic role
  labeling.
\newblock {\em Computational Linguistics\/} 34(2):257--287.

\bibitem[{Roth and Lapata(2016)}]{roth-lapata:2016:P16-1}
Michael Roth and Mirella Lapata. 2016.
\newblock Neural semantic role labeling with dependency path embeddings.
\newblock In {\em Proceedings of ACL\/}.

\bibitem[{Sennrich and Haddow(2016)}]{DBLP:conf/wmt/SennrichH16}
Rico Sennrich and Barry Haddow. 2016.
\newblock Linguistic input features improve neural machine translation.
\newblock In {\em Proceedings of WMT\/}.

\bibitem[{Socher et~al.(2013)Socher, Perelygin, Wu, Chuang, Manning, Ng, and
  Potts}]{socher-EtAl:2013:EMNLP}
Richard Socher, Alex Perelygin, Jean Wu, Jason Chuang, Christopher~D. Manning,
  Andrew Ng, and Christopher Potts. 2013.
\newblock Recursive deep models for semantic compositionality over a sentiment
  treebank.
\newblock In {\em Proceedings of EMNLP\/}.

\bibitem[{Sutskever et~al.(2014)Sutskever, Vinyals, and
  Le}]{DBLP:conf/nips/SutskeverVL14}
Ilya Sutskever, Oriol Vinyals, and Quoc~V. Le. 2014.
\newblock Sequence to sequence learning with neural networks.
\newblock In {\em Proceedings of NIPS\/}.

\bibitem[{Swayamdipta et~al.(2016)Swayamdipta, Ballesteros, Dyer, and
  Smith}]{DBLP:conf/conll/SwayamdiptaBDS16}
Swabha Swayamdipta, Miguel Ballesteros, Chris Dyer, and Noah~A. Smith. 2016.
\newblock Greedy, joint syntactic-semantic parsing with stack {LSTMs}.
\newblock In {\em Proceedings of CoNLL\/}.

\bibitem[{Thompson et~al.(2003)Thompson, Levy, and
  Manning}]{DBLP:conf/ecml/ThompsonLM03}
Cynthia~A. Thompson, Roger Levy, and Christopher~D. Manning. 2003.
\newblock A generative model for semantic role labeling.
\newblock In {\em Proceedings of ECML\/}.

\bibitem[{Titov et~al.(2009)Titov, Henderson, Merlo, and
  Musillo}]{TitovIjcai09}
Ivan Titov, James Henderson, Paola Merlo, and Gabriele Musillo. 2009.
\newblock Online projectivisation for synchronous parsing of semantic and
  syntactic dependencies.
\newblock In {\em Proceedings of IJCAI\/}.

\bibitem[{van~den Oord et~al.(2016)van~den Oord, Kalchbrenner, Espeholt,
  Kavukcuoglu, Vinyals, and Graves}]{DBLP:conf/nips/OordKEKVG16}
A{\"{a}}ron van~den Oord, Nal Kalchbrenner, Lasse Espeholt, Koray Kavukcuoglu,
  Oriol Vinyals, and Alex Graves. 2016.
\newblock Conditional image generation with {PixelCNN} decoders.
\newblock In {\em Proceedings of NIPS\/}.

\bibitem[{Zhao et~al.(2009)Zhao, Chen, Kazama, Uchimoto, and
  Torisawa}]{DBLP:conf/conll/ZhaoCKUT09}
Hai Zhao, Wenliang Chen, Jun'ichi Kazama, Kiyotaka Uchimoto, and Kentaro
  Torisawa. 2009.
\newblock Multilingual dependency learning: Exploiting rich features for
  tagging syntactic and semantic dependencies.
\newblock In {\em Proceedings of CoNLL\/}.

\bibitem[{Zhou and Xu(2015)}]{zhou-xu:2015:ACL-IJCNLP}
Jie Zhou and Wei Xu. 2015.
\newblock End-to-end learning of semantic role labeling using recurrent neural
  networks.
\newblock In {\em Proceedings of ACL\/}.

\end{thebibliography}
\bibliographystyle{emnlp_natbib}

\appendix

\newpage
\section{Hyperparameter values}
\label{sec:supplemental}

\begin{table}[h]
\centering
\begin{tabular}{@{\extracolsep{\fill}}l@{\hspace{6pt}}l@{\hspace{6pt}}c}
  \toprule
  &Semantic role labeler& \\
  \midrule
   & $d_w$ (word embeddings EN) & 100   \\
   & $d_w$ (word embeddings CH) & 128   \\
   & $d_{pos}$ (POS embeddings) &  16  \\
   & $d_l$ (lemma embeddings) & 100 \\
   & $d_h$ (LSTM hidden states) &   512  \\
   & $d_{r}$ (role representation) & 128  \\
   & $d'_{l}$ (output lemma representation) & 128  \\
   & $J$ (BiLSTM depth) & 3  \\
   & $K$ (GCN depth) & 1  \\
  & $\beta$ (edge dropout) & .3  \\
  
   & learning rate & .01  \\
\bottomrule
\end{tabular}
\caption{\label{tab:hyperparameters} Hyperparameter values. 
}
\end{table}

\end{document}